\documentclass[runningheads]{llncs}

\usepackage{accv}
\usepackage{accvabbrv}

\usepackage{graphicx}
\usepackage{booktabs}
\usepackage{pifont}
\usepackage{placeins}
\usepackage{wrapfig}
\usepackage{amsmath}
\usepackage{multirow}
\usepackage{needspace}
\usepackage{tablefootnote}
\usepackage{threeparttable}
\usepackage[table]{xcolor}
\newcommand{\redfn}[1]{\textsuperscript{\textcolor{red}{#1}}}

\usepackage{amsmath,amssymb}
\makeatletter
\renewcommand\paragraph{\@startsection{paragraph}{4}{\z@}%
  {0.8ex plus 0.2ex minus 0.1ex}%
  {-0.8em}%
  {\normalfont\normalsize\bfseries}}
\makeatother

\usepackage[accsupp]{axessibility}  

\usepackage[pagebackref,breaklinks,colorlinks,citecolor=accvblue]{hyperref}

\usepackage{orcidlink}

\begin{document}

\title{EndoPrior-GS: Dynamic Endoscopic Reconstruction with a Joint Texture Prior} 

\titlerunning{EndoPrior-GS}



\author{
Jiaqi Huang\inst{1}\orcidlink{0009-0002-6116-0772} \and
Shidong Wang\inst{1}\orcidlink{0000-0003-1023-1286}%
\thanks{Corresponding author.} \and
Tong Xin\inst{2}\orcidlink{0000-0001-5479-262X} \and
Kabita Adhikari\inst{1}\orcidlink{0000-0001-8726-6503}
}

\authorrunning{J. Huang et al.}

\institute{
School of Engineering, Newcastle University,\\
Newcastle upon Tyne NE1 7RU, UK\\
\email{\{J.huang57,shidong.wang,kabita.adhikari\}@ncl.ac.uk}
\and
School of Computing, Newcastle University,\\
Newcastle upon Tyne NE4 5TG, UK\\
\email{tong.xin@ncl.ac.uk}
}
\maketitle

\begin{abstract} 
Dynamic endoscopic reconstruction is fundamental to robotic surgery and computer-assisted interventions. 
While 3D Gaussian Splatting (3DGS) realises real-time rendering, its application to deformable intraoperative environments remains constrained by spurious geometry and varying illuminations.   
To address these limitations, we introduce EndoPrior-GS, a novel pipeline that explicitly couples frame-extracted vision heuristics and estimated depth maps. 
EndoPrior-GS derives a joint texture prior from a tool-filtered valid
tissue mask, a non-specular photometric filter, and anatomical structural salience, yielding a probability map that guides primitive initialisation and subsequent density control.
The prior is further extended to the temporal domain through a texture-aware term that dynamically weighs pairwise primitive contributions during training.
We conduct extensive experiments on benchmark datasets EndoNeRF and SCARED, and the obtained results show that our method EndoPrior-GS reduces Flow Error by 27.7\% and 25.8\% over the representative approaches while preserving competitive rendering quality and real-time rendering speed. Our project website is available at \url{https://jiaqi-huang-77.github.io/EndoPrior-GS/}. 
\end{abstract}

\section{Introduction}

\begin{figure}[t]
\centering
\hspace*{-0.015\linewidth}%
\includegraphics[width=1.03\linewidth]{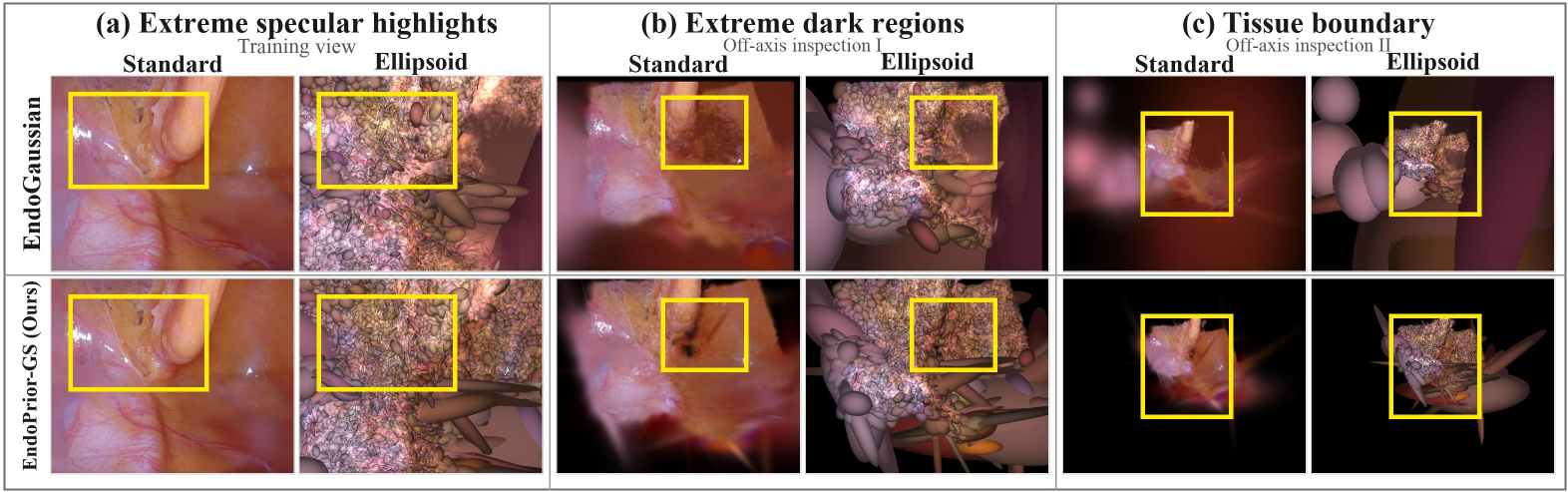}
\caption{
\textbf{Comparison of rendering results in challenging endoscopic regions.}  Panels (a)--(c) show extreme specular highlights, extreme dark
regions, and tissue boundaries, respectively.
Compared with EndoGaussian~\cite{endogaussian}, EndoPrior-GS (Ours) mitigates
semi-transparent floaters, oversized Gaussians near tissue
boundaries, and degenerate sheet-like Gaussians in low-texture
regions.
}
\label{fig:intro_qualitative}
\vspace{-18pt}
\end{figure}

In recent years, the development of computer-assisted interventions, such as diagnostic endoscopy, has shifted towards generating real-time rendering and high-fidelity 3D reconstruction of the tissue field to enhance intraoperative navigation and visual interpretation of complex, non-rigid anatomical structures ~\cite{3d_reconstruction_surgical,3d_reconstruction_surgical_train,3d_reconstruction_surgical_train_need,surgical_robots}. While continuously implicit representations such as Neural Radiance Fields (NeRF) have shown promise for this task, the considerable computational overhead has hindered true real-time rendering~\cite{nerf,endonerf}.    

The emergence of 3D Gaussian Splatting (3DGS) has fundamentally transformed novel view synthesis, enabling high-speed synthesis via either explicit point-based representations~\cite{3dgs,pixelgs,endogs} or pixel-wise re-projection~\cite{pixelsplat,splatter_image,mvsplat,depthsplat}, coupled with tile-based rasterisation. However, adapting vanilla 3DGS to dynamic endoscopic sequences is challenging. Standard 3DGS relies heavily on Structure-from-Motion (SfM) frameworks like COLMAP~\cite{sfm} for initial sparse point cloud generation, which usually fails in endoscopy because of the uniform, featureless textures of internal organs and the severe spatial constraints of monocular camera trajectories~\cite{endogaussian,free_surgs}. 

\begin{wrapfigure}{r}{0.45\linewidth}
\vspace{-37pt}
\centering
\includegraphics[width=\linewidth]{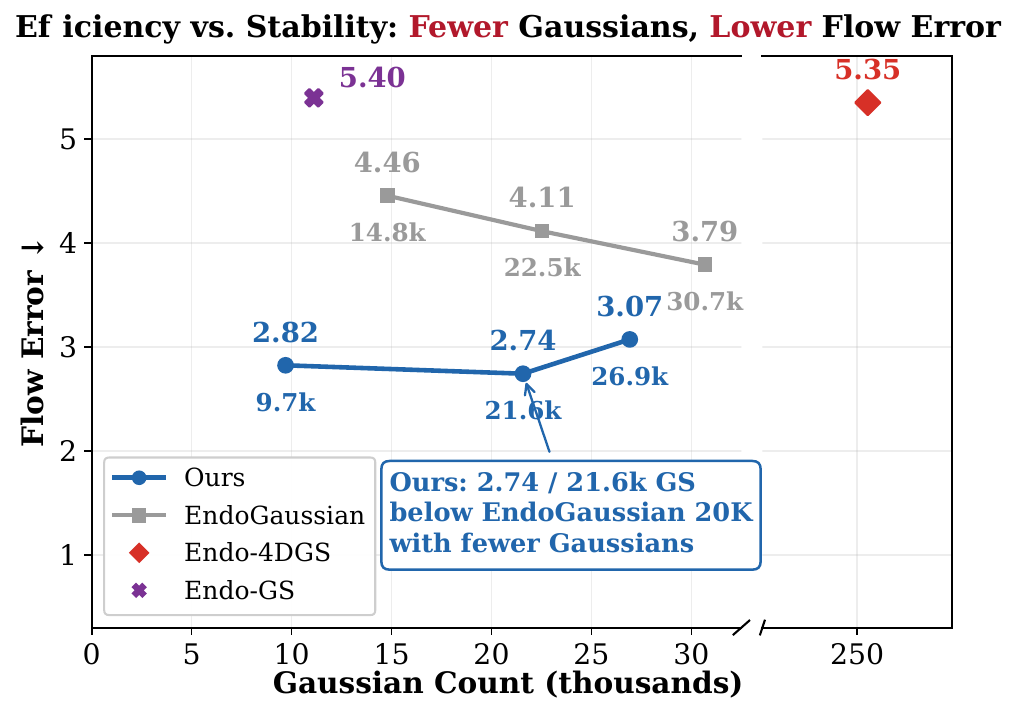}
\caption{
\textbf{Efficiency and Stability.}
EndoPrior-GS (Ours) achieves lower Flow Error with fewer Gaussians.
}
\label{fig:efficiency_stability}
\vspace{-22pt}
\end{wrapfigure}

The depth-reprojection pipelines~\cite{endogaussian,surgicalgaussian,endo_4dgs}, on the other hand, leverage monocular depth estimators to guide the initialisation of Gaussian primitives. Whilst computationally expedient, these approaches treat every pixel uniformly. Specifically, depth estimation cannot precisely capture geometry on highly reflective instruments, preserve sharp anatomical boundaries, and distinguish ambiguous regions under varying illumination as illustrated in Fig.~\ref{fig:intro_qualitative}. Furthermore, when dynamic instruments pass across the tissue bed, both point-based and depth reprojection-based approaches struggle to maintain long-range inter-frame consistency, as the temporal correspondence of underlying tissue becomes non-rigid and deformable. The Flow Error for benchmarked approaches is shown in Fig.~\ref{fig:efficiency_stability}.

To overcome these limitations, we present a novel pipeline, \texttt{EndoPrior-GS}, that explicitly couples the depth maps with frame-wise vision heuristics to guide splatting with structural texture information. Firstly, we construct a probability map by taking the product of a binary tool-filtered valid
tissue mask, a non-specular photometric filter, and anatomical structural salience. This probability map then serves as a filter and priority guide that can be formulated and accumulated into a persistent texture-prior score to dynamically regulate the entire splatting process. We further introduce a texture-aware temporal regularisation term that can be imposed on the original 3DGS optimisation pipeline to dynamically assign different strengths to local neighbourhood primitive relations to enforce inter-frame consistency.

With EndoGaussian~\cite{endogaussian} as baseline, our
contributions are summarised below:
\begin{itemize}
\item We propose a training-free \texttt{Joint Texture Prior}, derived from frame-wise visual heuristics, to provide an explicit prior-based mechanism for guiding Gaussian primitive evolution.

\item We develop a \texttt{Prior-Guided Initialisation and Density Control} strategy that regulates Gaussian initialisation and density updates, reducing redundant growth in geometrically distorted regions.

\item We introduce a \texttt{Texture-Aware Temporal Consistency} term that dynamically assigns different strengths to local neighbourhood primitive relations to enforce inter-frame consistency. 

\end{itemize}

We conducted experiments to show that EndoPrior-GS reduces Flow Error by 27.7\% and 25.8\% on EndoNeRF and SCARED, respectively, while improving temporal consistency and suppressing redundant and unstable Gaussian growth.

\section{Related Work}
\label{related work}
\subsection{Endoscopic Neural Reconstruction}
Dynamic endoscopic reconstruction is hindered by non-rigid tissue deformation, restricted camera trajectories, tool occlusion, specular reflections, motion blur, and ambiguous or weak tissue texture~\cite{endonerf,e_dssr,serv_ct,endogs}.
Neural rendering methods have adapted radiance-field and neural-surface representations to surgical scenes.
EndoNeRF~\cite{endonerf} reconstructs deformable tissues with neural rendering, mask-guided ray sampling, and stereo depth priors, while EndoSurf~\cite{endosurf} combines deformation, signed-distance, and radiance fields for dynamic surface reconstruction. 
These implicit formulations encode scene support as continuous fields queried along rays, requiring costly ray sampling and leaving no persistent local elements for direct refinement or explicit temporal constraints.

\subsection{Dynamic Endoscopic Reconstruction with 3DGS}

Explicit 3D Gaussian Splatting offers a more controllable alternative for efficient dynamic endoscopic reconstruction~\cite{3dgs}. Existing methods mainly incorporate auxiliary cues at specific stages.

Several methods address the initial Gaussian construction by  
re-projection from depth maps. EndoGaussian~\cite{endogaussian} proposes Holistic Gaussian Initialisation (HGI), where pixels from input frames are re-projected using estimated depth maps. SurgicalGaussian~\cite{surgicalgaussian} follows the same general direction by using depth priors. Endo-4DGS~\cite{endo_4dgs} uses pseudo-depth maps for monocular Gaussian initialisation and applies confidence-guided depth learning during optimisation. These methods mainly apply depth and mask cues to construct or supervise the initial Gaussian set; subsequent primitive growth and pruning remain governed by standard 3DGS optimisation criteria.

Other works improve endoscopic Gaussian deformation, surface geometry, or appearance modelling. 
Endo-GS~\cite{endogs} utilises depth-guided supervision and surface-aligned regularisation to improve deformable tissue reconstruction under tool occlusion. 
SurgicalGaussian~\cite{surgicalgaussian} imposes consistency losses on Gaussian position and covariance to regularise local motion. 
Surgical Gaussian Surfels~\cite{surgical_gaussian_surfels} models tissue geometry with Gaussian surfels to avoid unrestricted volumetric scaling of standard 3DGS.
Endo-4DGX~\cite{endo_4dgx} incorporates illumination embeddings for illumination-adaptive colour prediction. 
Although these methods are effective, they mostly address specific failure modes through targeted supervision, regularisation, geometric modelling, or appearance conditioning.


Rather than treating endoscopy-specific challenges merely as isolated terms in the training objective or as stand-alone correction modules, we re-examine the optimisation flow of 3DGS. The proposed EndoPrior-GS explicitly couples frame-extracted vision heuristics into the Gaussian splatting optimisation pipeline throughout training.

\section{Method}

\subsection{Overview}
\label{sec:method_overview}

\begin{figure}[h]
\vspace{-18pt}
\centering
\includegraphics[width=1.02\linewidth]{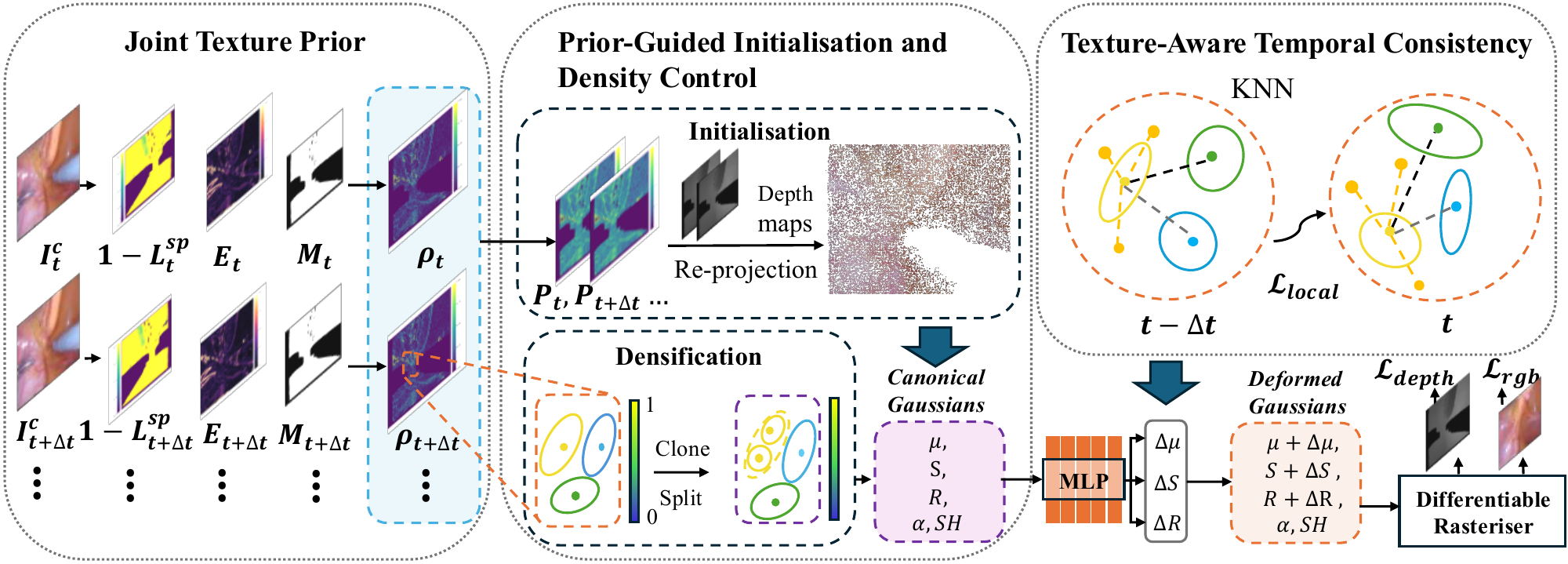}
\caption{
\textbf{The proposed EndoPrior-GS framework.} 
}
\label{fig:overview}
\vspace{-12pt}
\end{figure}

\label{sec:method_overview}

The core pipeline of the proposed \texttt{EndoPrior-GS} is illustrated in Fig.~\ref{fig:overview}. Specifically, we construct a unified texture prior that operates as a training-free heuristic mechanism. Computed directly from low-level image cues, this prior circumvents the need for heavy feature-extraction networks commonly found in standard approaches. The joint texture prior, alongside the raw input frame, is first normalised into an initialisation sampling distribution that adaptively determines whether a 3D Gaussian primitive should be initialised at a given pixel location. Furthermore, we formulate a texture-prior scoring function to dynamically govern the lifecycle of the Gaussians during the subsequent densification and pruning stages. Lastly, a texture-aware temporal regularisation term is integrated into the optimisation flow to enforce spatio-temporal consistency, ensuring stable and high-fidelity reconstruction of highly deformable dynamic endoscopic scenes.

\subsection{Joint Texture Prior}
\label{sec:3.2prior}

\begin{wrapfigure}{r}{0.46\linewidth}
    \vspace{-9mm}
    \centering
    \includegraphics[width=0.31\linewidth,trim=4 4 4 4,clip]{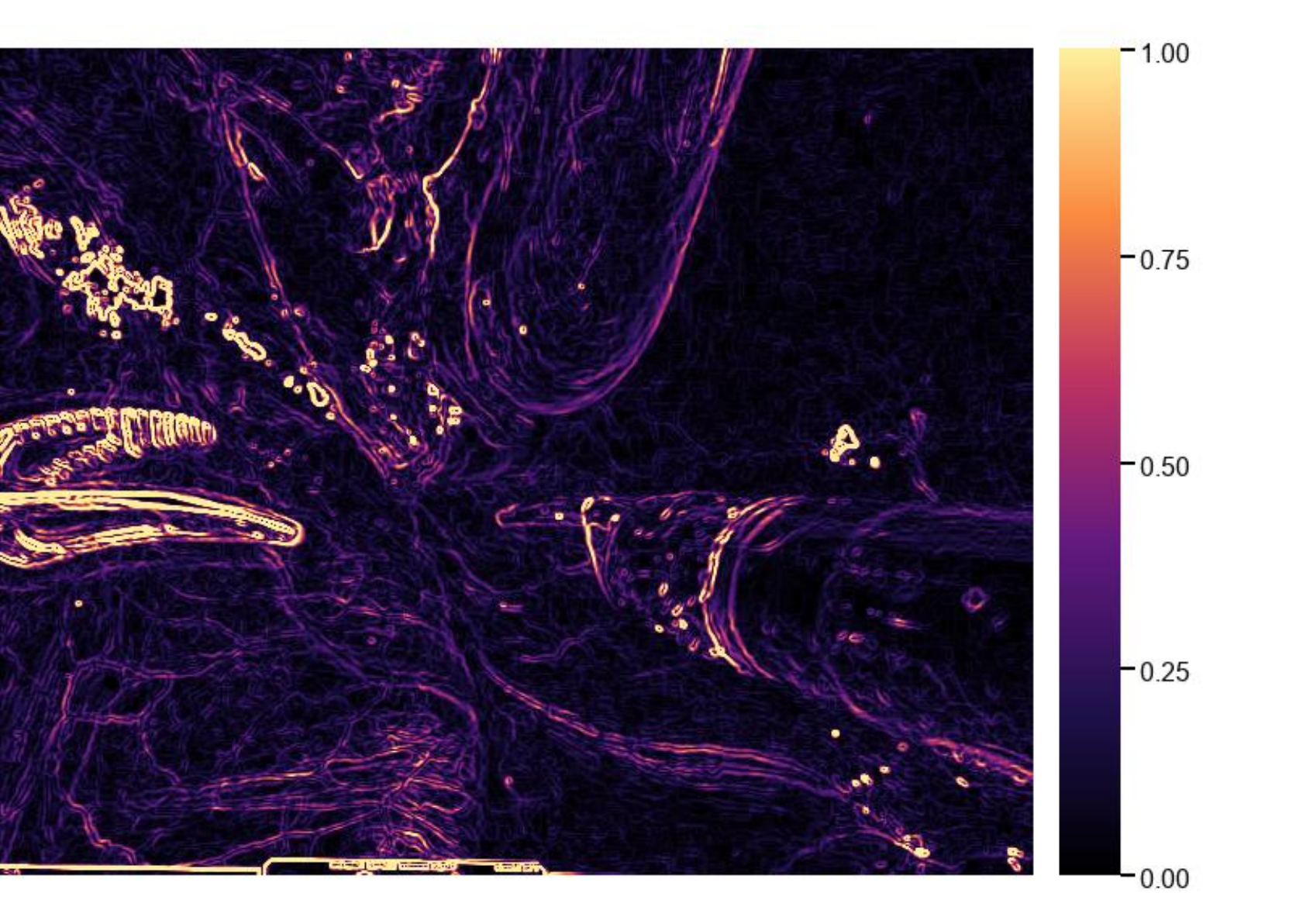}
    \includegraphics[width=0.31\linewidth,trim=4 4 4 4,clip]{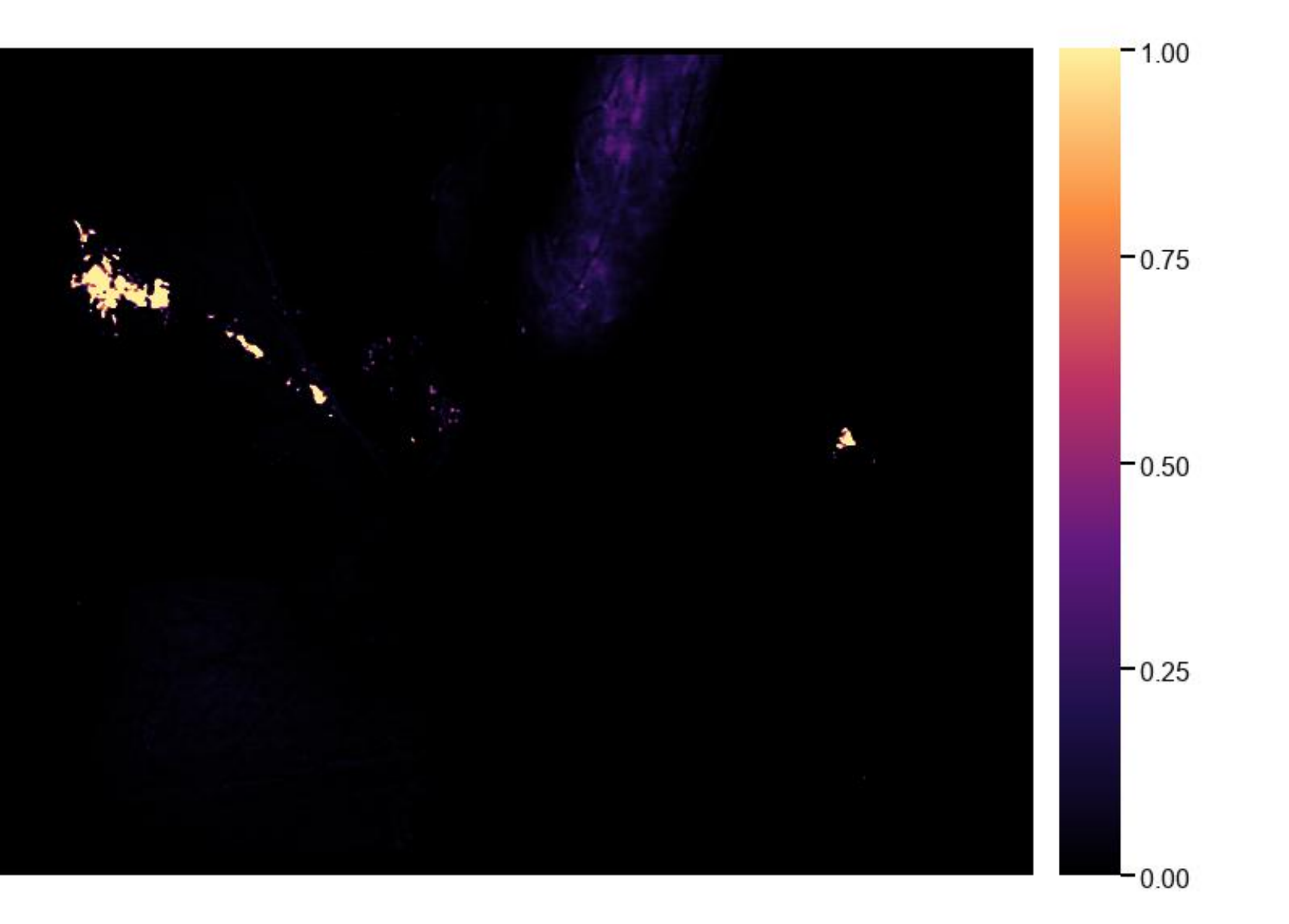}
    \includegraphics[width=0.31\linewidth,trim=4 4 4 4,clip]{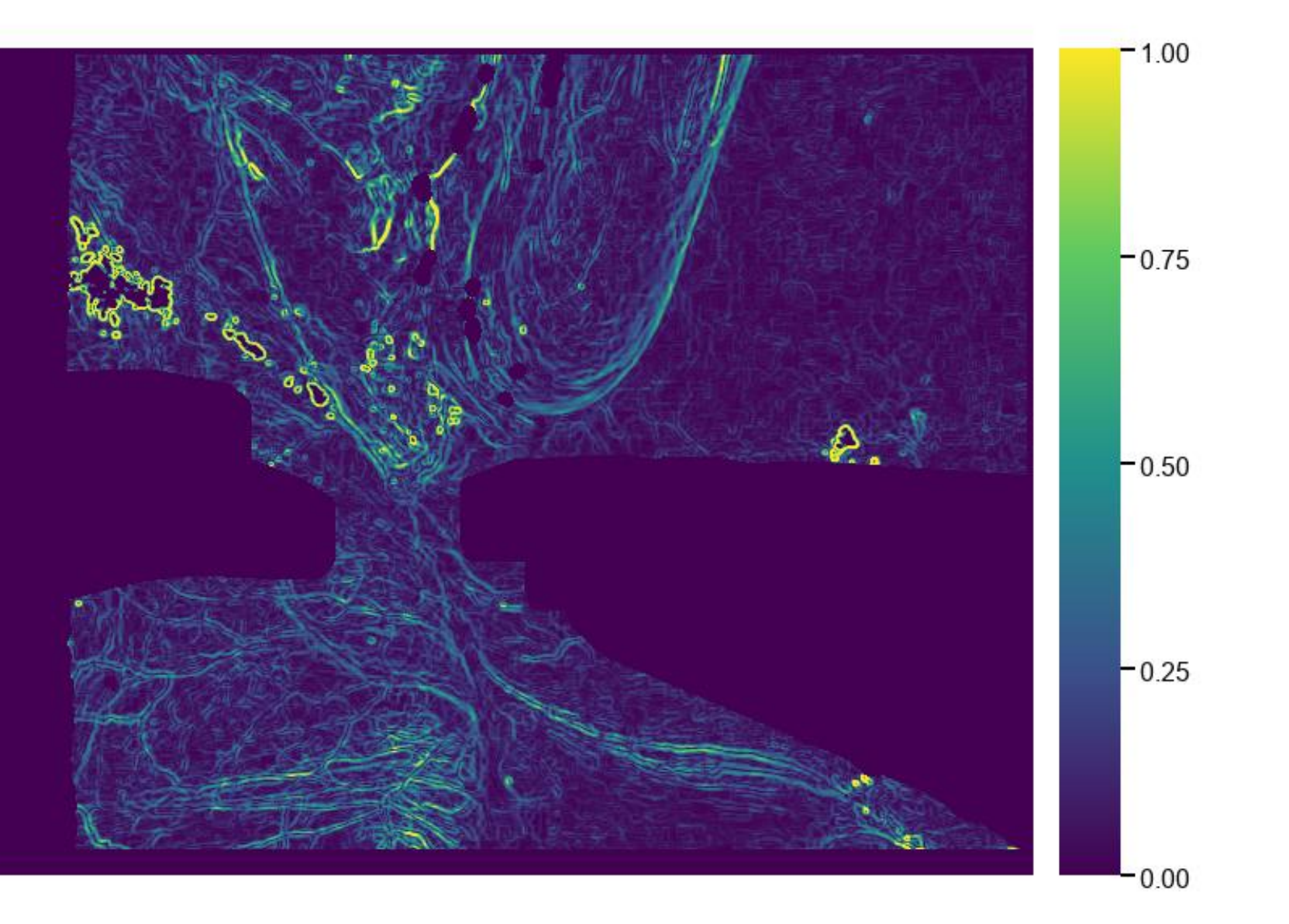}
    \vspace{-2mm}
    \caption{\textbf{Joint Texture Prior.}
    \(E_t\), \(L_t^{\mathrm{sp}}\), and \(\rho_t=M_t\odot(1-L_t^{\mathrm{sp}})\odot E_t\).}
    \label{fig:joint_texture_prior}
    \vspace{-10mm}
\end{wrapfigure}

We construct a joint texture prior to address the scenarios where standard 3DGS struggles when applied to dynamic endoscopic scenes, which are pervasively corrupted by instrument occlusions or background noise, specular reflections and lack of distinctive textures. These adverse factors often trigger floating artefacts, as illustrated in Fig.~\ref{fig:intro_qualitative}.

To mitigate this, we propose a joint texture prior $\rho_t$ as a pixel-wise spatial probability map over the image domain $\Omega_t$, which denotes the discrete 2D pixel-coordinate space of the image frame $t$. For any pixel location $\mathbf{u}=(u,v)\in\Omega_t$, $\rho_t(\mathbf{u})$ is computed directly from low-level image cues:
\begin{equation} 
\rho_t(\mathbf{u}) = M_t(\mathbf{u}) \bigl(1-L_t^{\mathrm{sp}}(\mathbf{u})\bigr) E_t(\mathbf{u}), \label{eq:pixel_reliability} 
\end{equation}
where \(M_t(\mathbf u)\in\{0,1\}\) is the value of the binary tool-filtered valid
tissue mask at pixel \(\mathbf u\), \(L_t^{sp}(\mathbf u)\in[0,1]\) is the
specular likelihood at \(\mathbf u\), and \(E_t(\mathbf u)\in[0,1]\) is the
anatomical structural salience at \(\mathbf u\).

Specifically, the mask \(M_t\) is extracted directly from the dataset, when \(M_t(\mathbf{u})=1\) indicates usable tissue and \(M_t(\mathbf{u})=0\) indicates tool or invalid pixels. The reflection intensity map \(L_t^{\mathrm{sp}}\) is estimated from RGB intensity and saturation cues\cite{hsv,specular}. The specular likelihood \(L_t^{\mathrm{sp}}(\mathbf{u})\in[0,1]\) at pixel \(\mathbf{u}\) is computed as:
\begin{equation}
L_t^{\mathrm{sp}}(\mathbf{u})
=
\max\left\{
\sigma\left(\frac{V_t(\mathbf{u})-\tau_v}{\gamma_v}\right)
\sigma\left(\frac{\tau_s-S_t(\mathbf{u})}{\gamma_s}\right),
C_t(\mathbf{u})
\right\},
\label{eq:specular_likelihood}
\end{equation}
where \(V_t(\mathbf{u})=\max_c I_t^c(\mathbf{u})\) is the maximum RGB intensity, \(I_t^c(\mathbf{u})\) is the intensity of RGB colour channel \(c\in\{R,G,B\}\) at pixel \(\mathbf{u}\) in frame \(t\), and \(S_t(\mathbf{u})=(V_t(\mathbf{u})-\min_c I_t^c(\mathbf{u}))/(V_t(\mathbf{u})+\epsilon)\) is the saturation computed from the RGB channels. \(C_t\) is a binary saturated-channel indicator map, and
\(C_t(\mathbf u)\in\{0,1\}\) is its value at pixel \(\mathbf u\). Specifically, \(C_t(\mathbf u)=1\) when at least two RGB channels at
\(\mathbf u\) exceed \(\tau_o\), and \(C_t(\mathbf u)=0\) otherwise. \(\sigma(\cdot)\) denotes the standard sigmoid activation function, which converts the intensity and saturation response into the continuous interval \([0, 1]\); \(\tau_v\) and \(\tau_s\) are the brightness and saturation thresholds, \(\gamma_v\) and \(\gamma_s\) are the temperature parameters controlling the sigmoid transition around \(\tau_v\) and \(\tau_s\).



Concurrently, the structural salience value \(E_t(\mathbf{u})\in[0,1]\) measures the strength
of anatomical edges and texture at pixel \(\mathbf{u}\), which provides useful
landmarks for tracking non-rigid tissue deformation. To compute this, the RGB frame is converted to a greyscale intensity map \(I_t\). The horizontal and vertical gradients \(G_x(\mathbf{u})\) and \(G_y(\mathbf{u})\) at pixel \(\mathbf{u}\) are calculated by convolving the image with directional \(3 \times 3\) Sobel kernels, \(K_{x}\) and \(K_{y}\):
\begin{equation}
G_x(\mathbf{u}) = (K_x \ast I_t)(\mathbf{u}), \quad G_y(\mathbf{u}) = (K_y \ast I_t)(\mathbf{u}),
\end{equation}
where the kernels are defined as:
\begin{equation}
K_x = \begin{bmatrix} -1 & 0 & 1 \\ -2 & 0 & 2 \\ -1 & 0 & 1 \end{bmatrix}, \quad K_y = \begin{bmatrix} -1 & -2 & -1 \\ 0 & 0 & 0 \\ 1 & 2 & 1 \end{bmatrix}.
\end{equation}

The unnormalised edge magnitude is obtained via the Euclidean norm \(\hat{E}_t(\mathbf{u}) = \sqrt{G_x(\mathbf{u})^2 + G_y(\mathbf{u})^2}\). To bound this structural prior to the continuous range \([0, 1]\), 
we normalise the edge magnitude using a robust percentile computed
over valid-tissue pixels within the image domain $\Omega_t$:

\begin{equation}
E_t(\mathbf{u})
=
\min\left\{
\frac{\hat{E}_t(\mathbf{u})}
{Q_p\!\left(\{\hat{E}_t(\mathbf{x}) \mid \mathbf{x}\in\Omega_t,\ M_t(\mathbf{x})=1\}\right)+\epsilon},
1
\right\},
\label{eq:e_t}
\end{equation}

where \(Q_p(\cdot)\) is the \(p\)-th percentile, and \(\epsilon \) is a small numerical stabiliser. The cue maps and the resulting joint texture prior are illustrated in Fig.~\ref{fig:joint_texture_prior}.

\subsection{Prior-Guided Initialisation and Density Control}
\label{sec:3.3density_control}
The joint texture prior is explicitly leveraged to guide the initialisation, densification, and pruning of Gaussian primitives. This approach fundamentally departs from the standard adaptive density-control operations of 3DGS~\cite{3dgs}, ensuring that the Gaussian allocation evolves selectively and aligns with the extracted structurally valid tissue regions.

\paragraph{Initialisation:} 
Standard 3DGS initialises primitives using sparse Structure-from-Motion (SfM)~\cite{sfm} point clouds. In endoscopic scenes, this triggers catastrophic failures as points erroneously anchor to transient surgical-tool regions or shifting specular highlights. To overcome this, 
we leverage the joint texture prior \(\rho_t\) to construct an initialisation sampling
distribution \(P_t\) over the image domain \(\Omega_t\) at frame \(t\). For each pixel \(\mathbf{u}\in\Omega_t\), its sampling probability is defined as:
\begin{equation}
P_t(\mathbf{u}) =
\eta_{\mathrm{app}}
\frac{\widetilde{M}_t(\mathbf{u})}
{\sum_{\mathbf{x}\in\Omega_t}\widetilde{M}_t(\mathbf{x})+\epsilon}
+
(1-\eta_{\mathrm{app}})
\frac{\rho_t(\mathbf{u})}
{\sum_{\mathbf{x}\in\Omega_t}\rho_t(\mathbf{x})+\epsilon},
\label{eq:init_sampling}
\end{equation}
where \(\epsilon \) is a small numerical stabiliser, and \(\eta_{\mathrm{app}}\in[0,1]\) is a balancing coefficient. The first term enforces uniform coverage over the valid tissue, whilst the second term prioritises non-specular, structurally rich regions. To prevent initialisation on unstable tool boundaries, the raw mask \(M_{t}\) is conservatively refined via morphological erosion:
$\widetilde{M}_t = M_t \ominus B_{k}, \quad \text{where} \quad \widetilde{M}_t(\mathbf{u}) = \min_{\mathbf{v}\in B_{k}} M_t(\mathbf{u}+\mathbf{v})$, with \(\ominus \) denoting the erosion operator and \(B_{k}\) representing an elliptical structuring element of size \(k \times k\) (\(k = 9\)).

After sampling a set of initialisation pixels from \(P_t\), we back-project them with the
estimated depth maps to form the initial Gaussian point cloud.
\paragraph{Densification:} 

The complex and dynamic nature of endoscopic imaging, including specular highlights, tool boundaries, or motion blur, poses significant challenges for primitive densification, often resulting in severe geometric distortions. We address these vulnerabilities by introducing a persistent texture-prior score \(\bar s_i\) for each Gaussian that samples \(\rho_t\) at its projected image coordinates across frames, which attenuates densification for primitives repeatedly projected onto low-prior regions.

To obtain the persistent texture-prior score \(\bar{s}_i\), we first project each deformed Gaussian centre denoted by \(\boldsymbol{\mu}_{i,t}\in\mathbb{R}^3\) at frame \(t\) onto the image plane as \(\mathbf{c}_{i,t}=\pi(\boldsymbol{\mu}_{i,t};\Pi_t)\), where \(\pi(\cdot;\Pi_t)\) is the camera projection function with camera parameters \(\Pi_t\), and \(\mathbf{c}_{i,t}\in\mathbb{R}^2\) is the projected image coordinate. 

The texture-prior value of primitive \(i\) at frame \(t\) is computed as \(s_{i,t}= m_{i,t}\,\mathcal B(\rho_t,\mathbf c_{i,t})\), where \(\mathcal{B}\) denotes bilinear interpolation of \(\rho_t\) at \(\mathbf{c}_{i,t}\), and \(m_{i,t}\in\{0,1\}\) is the projection-validity indicator that is set to \(m_{i,t}=1\) 
when $\mathbf{c}_{i,t}$ lies within the image bounds and
$M_t(\mathbf{c}_{i,t})=1$, and to $m_{i,t}=0$ otherwise.
The texture-prior score \(\bar s_i\) is then obtained by averaging valid samples:
\begin{equation} 
\bar s_i = \frac{\sum_t s_{i,t}} {\sum_t m_{i,t}+\epsilon}.
\label{eq:primitive_texture_prior_score} 
\end{equation}

\begin{wrapfigure}[12]{r}{0.48\columnwidth}
\vspace{-10mm}
\centering
\includegraphics[width=0.48\textwidth]{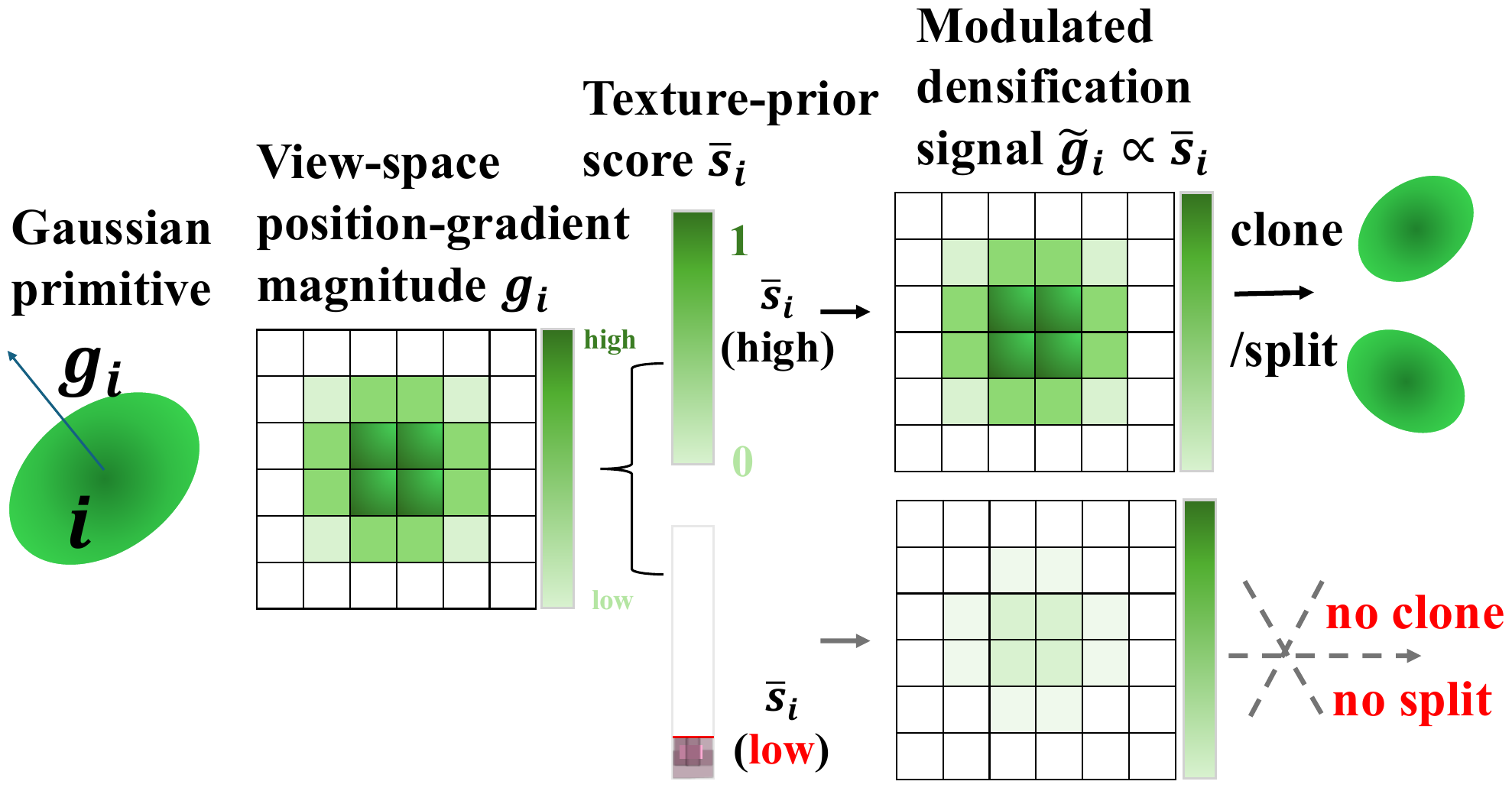}
\caption{
\textbf{Prior-Guided Densification.} The densification trigger is modulated by the texture-prior score \(\bar s_i\).
}
\label{fig:densification}
\end{wrapfigure}


The densification, therefore, can be modulated by incorporating the texture-prior score \(\bar s_i\) derived from Eq. (\ref{eq:primitive_texture_prior_score}), and denoted as:
\begin{equation}
\tilde g_i
=
\left[
\lambda_{\min}^{\mathrm{den}}
+
(1-\lambda_{\min}^{\mathrm{den}})\bar s_i
\right]g_i ,
\label{eq:densification}
\end{equation}
where \(g_i\) is the view-space position-gradient magnitude, and \(\lambda_{\min}^{\mathrm{den}}\in[0,1]\) is the minimum retained densification weight. This modulation attenuates densification triggers from artefact-induced gradients, thereby reducing redundant Gaussian growth around geometrically distorted regions. This modulated densification mechanism is illustrated in Fig.~\ref{fig:densification}, and more visuals can be found in Fig.~\ref{fig:allocation_analysis}.

\paragraph{Pruning:}


Our pruning strategy explicitly integrates the obtained texture-prior score \(\bar s_{i}\) alongside standard density-control criteria. Concretely, a primitive needs to satisfy the following criteria: \( n_i \ge n_{\min}, \qquad \bar s_i < \tau_{\mathrm{prior}}, \qquad \bar\alpha_i < \tau_\alpha\), where \(n_i=\sum_t m_{i,t}\) denotes the valid-projection count of Gaussian primitive \(i\), \(\bar\alpha_i\) is the running mean opacity during training, \(\tau_{\mathrm{prior}}\) and \(\tau_\alpha\) are the primitive texture-prior threshold and the opacity threshold. This criterion targets semi-transparent primitives that are repeatedly projected onto low-prior regions, while avoiding aggressive pruning of newly created or insufficiently sampled Gaussians.

\subsection{Texture-Aware Temporal Consistency}
\label{sec:reg}

While prior-guided initialisation and density control optimise the spatial distribution of Gaussian primitives, spatially heterogeneous anatomical motion may drive the deformation field to produce incoherent local offsets. 
To address this limitation, we incorporate a texture-aware temporal regularisation term that dynamically adjusts each primitive's contribution over time and encourages neighbouring Gaussian primitives to move coherently.

\begin{wrapfigure}[10]{r}{0.48\linewidth}
    \centering
    \vspace{-9mm}
    \includegraphics[width=\linewidth]{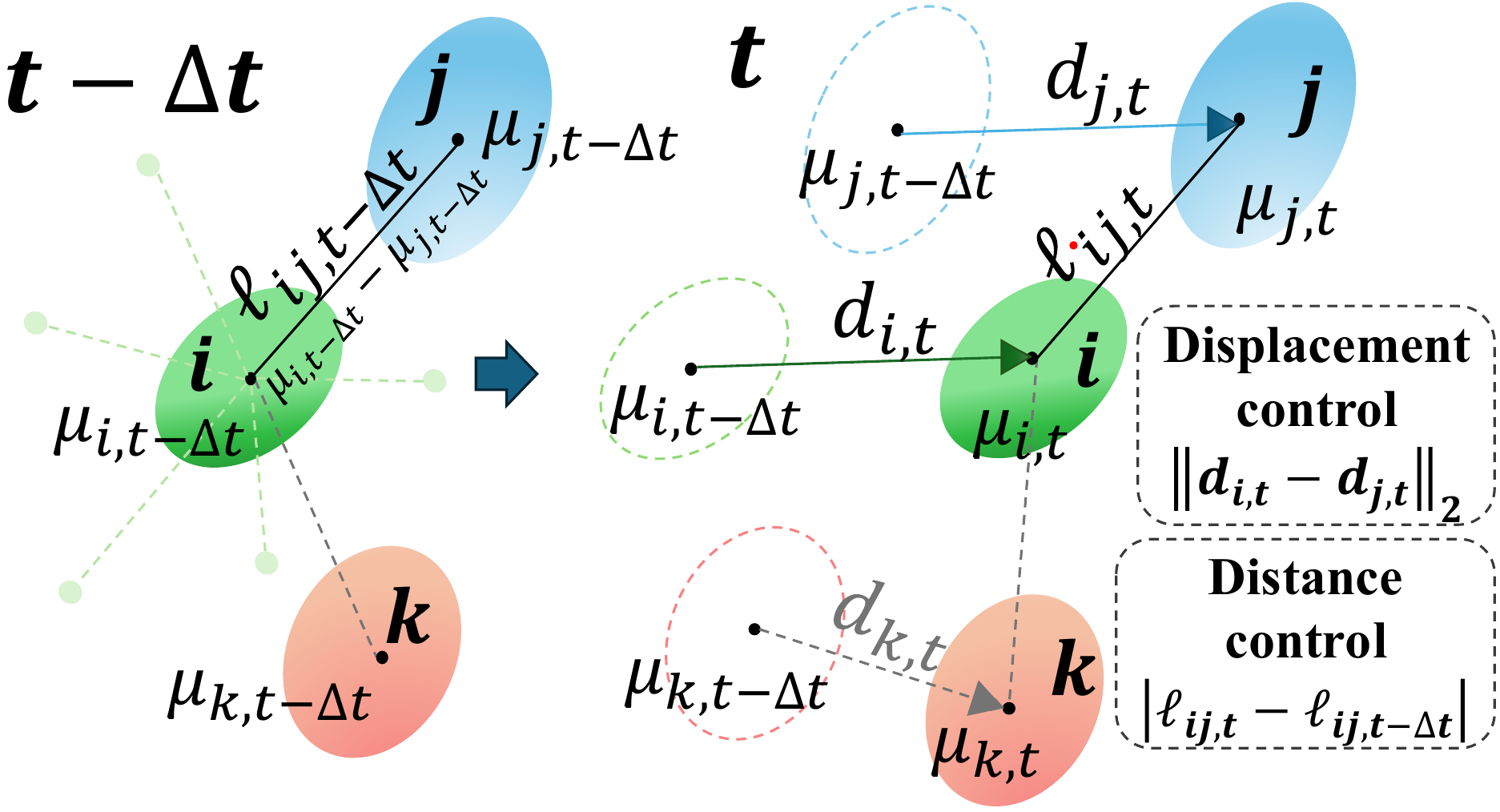}
    \caption{\textbf{Temporal Regularisation.}}
    \label{fig:motion_reg}
\end{wrapfigure}

Since all frames share a canonical Gaussian set, each primitive has a consistent identity. At frame \(t\), we first construct an eligible primitive set \(\mathcal{S}_{t}\) containing primitives that have valid projections at both \(t-\Delta t\) and \(t\), and meet the criteria of having a minimum number of valid projected samples, such that \(n_i \ge n_{\min}\). Here, \(\Delta t\) represents the normalised interval between adjacent training frames, and \(n_{i}\) denotes the valid-projection count of primitive \(i\). Then, for each eligible primitive \(i \in \mathcal{S}_t\), we assign a primitive-specific motion weight \(w_{i}^{\mathrm{mot}}\) derived from its texture-prior score \(\bar s_i\), which is formulated as:
\begin{equation}
w_i^{\mathrm{mot}}
=
\sigma\left(
\frac{\bar s_i-\tau_m}{\gamma_m}
\right),
\label{eq:motion_weight}
\end{equation}
where \(\sigma(\cdot)\) denotes the standard sigmoid activation function that maps the resulting scalar to the continuous interval \([0, 1]\), \(\tau _{m}\) is a texture-prior threshold for motion weight, and \(\gamma _{m}\) represents a temperature scaling parameter.

We further extend this to pairwise primitive relations to ensure local deformation constraints are confined to pairs where both primitives are mutually trustworthy. Specifically, we compute a pairwise spatio-temporal affinity weight \(w_{ij,t}\) between primitive \(i\) and its neighbour \(j\) at frame \(t\), which is defined as:
\begin{equation}
w_{ij,t}
=
\exp\left(
-
\frac{
\left\|
\boldsymbol{\mu}_{i,t-\Delta t}
-
\boldsymbol{\mu}_{j,t-\Delta t}
\right\|_2^2
}{
2\sigma_{\mathrm{loc}}^2
}
\right)
w_i^{\mathrm{mot}}w_j^{\mathrm{mot}},
\label{eq:local_affinity}
\end{equation}
where \(\sigma_{\mathrm{loc}}\) controls the spatial range of the local
neighbourhood, and \(\boldsymbol{\mu}_{i,t-\Delta t}\) and \(\boldsymbol{\mu}_{j,t-\Delta t}\) denote the 3D spatial centres of primitives \(i\) and \(j\) at the preceding reference frame \(t-\Delta t\), respectively. 

The proposed texture-aware temporal consistency loss that leverages the above prior-driven weights to ensure a smooth deformation and maintain structural coherence across neighbouring frames is specifically defined as:

\begin{equation}
\begin{aligned}
\mathcal{L}_{\mathrm{local}}
=
&
\lambda_{\mathrm{disp}}\frac{
\sum_{i\in\mathcal{S}_t}
\sum_{j\in\mathcal{N}_K^{(t)}(i)}
w_{ij,t}
\left\|
\mathbf{d}_{i,t}
-
\mathbf{d}_{j,t}
\right\|_2
}{
\sum_{i\in\mathcal{S}_t}
\sum_{j\in\mathcal{N}_K^{(t)}(i)}
w_{ij,t}
+
\epsilon
}
\\
&
+
\lambda_{\mathrm{dist}}
\frac{
\sum_{i\in\mathcal{S}_t}
\sum_{j\in\mathcal{N}_K^{(t)}(i)}
w_{ij,t}
h_{\delta}
\left(
\left|
\ell_{ij,t}
-
\ell_{ij,t-\Delta t}
\right|
\right)
}{
\sum_{i\in\mathcal{S}_t}
\sum_{j\in\mathcal{N}_K^{(t)}(i)}
w_{ij,t}
+
\epsilon
},
\end{aligned}
\label{eq:local_motion_loss}
\end{equation}
where \(\mathcal{N}^{(t)}_K(i)\) is the local neighbourhood of primitive \(i\), containing the \(K\) nearest Gaussian primitives in \(\mathcal{S}_t\setminus\{i\}\) with  distances measured at the preceding reference frame \(t-\Delta t\). The adjacent training frame displacements of primitives \(i\) and \(j\) are
\(\mathbf{d}_{i,t}=\boldsymbol{\mu}_{i,t}-\boldsymbol{\mu}_{i,t-\Delta t}\)
and
\(\mathbf{d}_{j,t}=\boldsymbol{\mu}_{j,t}-\boldsymbol{\mu}_{j,t-\Delta t}\), respectively. The pairwise Gaussian centre distance of primitive \(i\) and \(j\) at frame \(t\) is defined as \(\ell_{ij,t}=\left\|\boldsymbol{\mu}_{i,t}-\boldsymbol{\mu}_{j,t}\right\|_2\). The function \(h_{\delta}(\cdot)\) denotes the Huber penalty with transition parameter \(\delta\)~\cite{huber}, \(\lambda_{\mathrm{disp}}\) and \(\lambda_{\mathrm{dist}}\) balance the first and second terms, and \(\epsilon\) is a small constant for numerical stability. The first term reduces isolated primitive motion drift by encouraging neighbouring primitives to have similar adjacent-frame displacements. The second term
discourages sudden local stretching or collapse by penalising changes in
pairwise Gaussian centre distances. The mechanism is illustrated in Fig.~\ref{fig:motion_reg}.

The training objective of our framework is formulated as a linear combination of baseline objectives and our proposed texture-aware constraint. Formally, 
\begin{equation}
\mathcal{L}_{\mathrm{total}}
=
\mathcal{L}_{\mathrm{rgb}}
+
\lambda_{\mathrm{depth}}\mathcal{L}_{\mathrm{depth}}
+
\lambda_{\mathrm{tv}}\mathcal{L}_{\mathrm{tv}}
+
\lambda_{\mathrm{local}}\mathcal{L}_{\mathrm{local}},
\label{eq:final_training_objective}
\end{equation}
where \(\mathcal{L}_{\mathrm{rgb}}\), \(\mathcal{L}_{\mathrm{depth}}\), and \(\mathcal{L}_{\mathrm{tv}}\) represent the masked RGB colour loss, depth loss, and total variation (TV) loss introduced by EndoGaussian~\cite{endogaussian}, respectively; the scaling coefficients \(\lambda _{\mathrm{depth}}\), \(\lambda _{\mathrm{tv}}\), and \(\lambda _{\mathrm{local}}\) are hyperparameters that balance the contribution of each respective penalty during the joint optimisation flow.

\section{Experiments}

\subsection{Experimental Setup}
\label{sec:experimental_setup}
\paragraph{Datasets.}
We evaluate our method on two publicly released dynamic endoscopic reconstruction benchmarks: \textbf{EndoNeRF}~\cite{endonerf} and \textbf{SCARED}~\cite{scared}.

The \textbf{EndoNeRF} dataset~\cite{endonerf} consists of six stereo video sequences captured from DaVinci robotic surgical procedures, totalling 807 frames. Each sequence has a resolution of $512 \times 640$ and exhibits significant non-rigid tissue deformation and tool occlusion. Following prior work~\cite{endo_4dgs,endogaussian}, we evaluate on two public cases (\textit{pulling} and \textit{cutting}) using a 7:1 train-test split.

The \textbf{SCARED} dataset~\cite{scared} consists of stereo RGB-D endoscopic sequences captured from porcine cadaver abdominal anatomies using a DaVinci endoscope. We evaluate on datasets 1, 2, 3, 6, and 7.  Following the standard protocol used in previous endoscopic reconstruction works~\cite{endogaussian,endosurf}, we use a fixed temporal interval split on each sequence into training and testing frames with a 7:1 ratio.

\paragraph{Metrics.}
We assess rendering quality using standard image metrics: Peak Signal-to-Noise Ratio (PSNR), Structural Similarity Index Measure (SSIM)~\cite{ssim}, and Learned Perceptual Image Patch Similarity (LPIPS)~\cite{lpips}. Depth accuracy is measured using root mean square error (RMSE) over pixels with valid depth annotations. To evaluate temporal consistency, we report a RAFT-based Flow Error~\cite{raft} between the rendered and ground-truth video sequences within valid tissue regions. We also report the Gaussian primitive count \(N_{\mathrm{GS}}\) after training and rendering speed in frames per second (FPS) to measure rendering efficiency. 


A rendered-depth temporal instability diagnostic is reported in Sec.~\ref{sec:temporal_consistency} to quantify adjacent-frame depth variation. The diagnostic is computed for an adjacent frame pair \((t-\Delta t,t)\) as \(\mathcal{I}_{t} = \frac{1}{|\Omega_{t-\Delta t,t}^{\mathrm{val}}|} \sum_{\mathbf{u}\in\Omega_{t-\Delta t,t}^{\mathrm{val}}} \left| \hat D_t(\mathbf{u}) - \hat D_{t-\Delta t}(\mathbf{u}) \right|\), where \(\hat D_t(\mathbf{u})\) denotes the rendered depth in frame \(t\) at pixel \(\mathbf{u}\). 
The evaluation domain is \(\Omega_{t-\Delta t,t}^{\mathrm{val}}=\left\{\mathbf{u}\in \Omega_t\;\middle|\;M_{t-\Delta t}(\mathbf{u})=M_t(\mathbf{u})=1,\; \hat D_{t-\Delta t}(\mathbf{u})\hat D_t(\mathbf{u})>0\right\}\), where \(M_{t-\Delta t}(\mathbf{u})=M_t(\mathbf{u})=1\) indicates that pixel \(\mathbf{u}\) is retained by the binary tool-filtered valid tissue masks in both frames, and \(\hat D_{t-\Delta t}(\mathbf{u})\hat D_t(\mathbf{u})>0\) indicates valid rendered depth values in both frames.

\paragraph{Implementation details.}

All experiments follow the training schedule as EndoGaussian~\cite{endogaussian}, with 1,000 coarse-stage iterations and 3,000 fine-stage iterations, and are conducted on an NVIDIA RTX 6000 Ada Generation GPU.

In Sec.~\ref{sec:3.2prior}, the specular likelihood in Eq.~\ref{eq:specular_likelihood} uses \(\tau_v=0.85\), \(\tau_s=0.35\), \(\gamma_v=0.04\), \(\gamma_s=0.08\), and saturated-channel threshold $\tau_o=0.92\). For Eq.~\ref{eq:e_t}, we set percentile \(p=99\).
In Sec.~\ref{sec:3.3density_control}, we set the initialisation balancing coefficient in Eq.~\ref{eq:init_sampling} to \(\eta_{\mathrm{app}}=0.45\). The texture-prior score \(\bar{s}_i\) is updated every 100 iterations, following 3DGS~\cite{3dgs}. Prior-guided densification applies Eq.~\ref{eq:densification} with \(\lambda^{\mathrm{den}}_{\min}=0.35\). During pruning, we set \(\tau_{\mathrm{prior}}=0.2\), \(n_{\min}=20\), opacity threshold \(\tau_{\alpha}=0.0075\). 
In Sec.~\ref{sec:reg}, the weight in Eq.~\ref{eq:motion_weight} is initialised with \(\tau_m=0.45\) and \(\gamma_m=0.08\). We set \(\lambda_{\mathrm{dist}}=0.2\), \(\lambda_{\mathrm{disp}}=1.0\), \(K=8\) and Huber transition parameter \(\delta=0.01\) in Eq.~\ref{eq:local_motion_loss}, respectively. The texture-aware temporal regularisation term is weighted by \(\lambda_{\mathrm{local}}=10^{-6}\) in Eq.~\ref{eq:final_training_objective}. 
We use fixed hyperparameters across all sequences.

\subsection{Comparison with Representative Methods}
\label{sota_compare}

\begin{figure}[h]
\vspace{-20pt}
    \centering
    \includegraphics[width=1.0\linewidth]{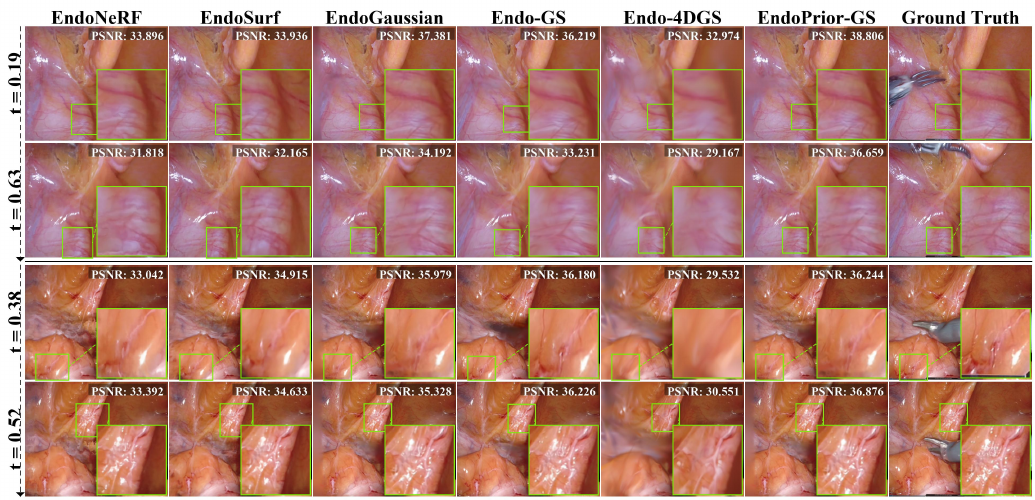}
    \caption{
    \textbf{Qualitative comparison with representative endoscopic reconstruction methods.} Results are shown on selected \emph{pulling} and \emph{cutting} sequences. Rows denote timestamps and columns denote methods or ground truth. 
    }
    \label{fig:final_result}
    \vspace{-15pt}
\end{figure}

\begin{table*}[h]
\centering
\scriptsize
\setlength{\tabcolsep}{4.2pt}
\renewcommand{\arraystretch}{1.08}
\definecolor{paperrefgrey}{RGB}{242,242,242}
\newcommand{\paperref}[1]{\cellcolor{paperrefgrey}{#1}}
\newcommand{\best}[1]{\cellcolor{red!18}\textbf{#1}}
\newcommand{\second}[1]{\cellcolor{orange!22}\textbf{#1}}
\newcommand{\dash}{--}
\begin{tabular}{lcccccc}
\toprule
Method
& PSNR$\uparrow$
& SSIM$\uparrow$
& LPIPS$\downarrow$
& RMSE$\downarrow$
& Flow Error$\downarrow$
& FPS$\uparrow$ \\
\midrule
\multicolumn{7}{c}{\textbf{EndoNeRF}} \\
\midrule
EndoNeRF~\cite{endonerf}\redfn{1}
& \paperref{32.039} & \paperref{0.925} & \paperref{0.081} & \paperref{\dash} & \paperref{\dash} & \paperref{\dash} \\
{}
& 27.519 & 0.933 & 0.128 & 2.522 & 5.421 & 0.14 \\
\cmidrule(lr){1-7}
EndoSurf~\cite{endosurf}\redfn{2}
& \paperref{34.992} & \paperref{0.954} & \paperref{0.113} & \paperref{1.000} & \paperref{\dash} & \paperref{\dash} \\
{}
& 35.004 & 0.845 & 0.243 & 2.584 & 5.284 & 12.84 \\
\cmidrule(lr){1-7}
EndoGaussian~\cite{endogaussian}\redfn{5}
& \paperref{37.849} & \paperref{0.963} & \paperref{0.054} & \paperref{\dash} & \paperref{\dash} & \paperref{\dash} \\
{}
& \second{37.322} & \second{0.958} & \best{0.060} & 2.482 & \second{3.791} & \second{279} \\
\cmidrule(lr){1-7}
Endo-GS~\cite{endogs}\redfn{3}
& \paperref{37.935} & \paperref{0.966} & \paperref{0.034} & \paperref{\dash} & \paperref{\dash} & \paperref{\dash} \\
{}
& 36.710 & 0.955 & \second{0.067} & \second{2.477} & 5.396 & 76 \\
\cmidrule(lr){1-7}
Endo-4DGS~\cite{endo_4dgs}\redfn{4}
& \paperref{37.205} & \paperref{0.957} & \paperref{0.038} & \paperref{\dash} & \paperref{\dash} & \paperref{\dash} \\
{}
& 35.353 & 0.924 & 0.104 & 2.686 & 5.369 & 15 \\
\cmidrule(lr){1-7}

\textbf{EndoPrior-GS (Ours)}
& \best{38.071} & \best{0.967} & 0.071
& \best{2.260} & \best{2.742} & \best{305} \\
\midrule
\multicolumn{7}{c}{\textbf{SCARED}} \\
\midrule
EndoNeRF~\cite{endonerf}\redfn{1}
& \paperref{\dash} & \paperref{\dash} & \paperref{\dash} & \paperref{\dash} & \paperref{\dash} & \paperref{\dash} \\
{}
& 24.345 & 0.768 & 0.313 & 2.723 & 5.894 & 0.06 \\
\cmidrule(lr){1-7}
EndoSurf~\cite{endosurf}\redfn{2}
& \paperref{23.637} & \paperref{0.794} & \paperref{0.351} & \paperref{\dash} & \paperref{\dash} & \paperref{\dash} \\
{}
& 25.020 & 0.802 & 0.356 & \second{2.501} & 5.401 & 2 \\
\cmidrule(lr){1-7}
EndoGaussian~\cite{endogaussian}\redfn{5}
& \paperref{27.042} & \paperref{0.827} & \paperref{0.267} & \paperref{\dash} & \paperref{\dash} & \paperref{\dash} \\
{}
& 27.070 & 0.744 & 0.249 & 2.675 & 5.232 & \second{269} \\
\cmidrule(lr){1-7}
Endo-GS~\cite{endogs}\redfn{3}
& \paperref{\dash} & \paperref{\dash} & \paperref{\dash} & \paperref{\dash} & \paperref{\dash} & \paperref{\dash} \\
{}
& 27.786 & 0.781 & 0.245 & 2.548 & \second{4.867} & 76 \\
\cmidrule(lr){1-7}
Endo-4DGS~\cite{endo_4dgs}\redfn{4}
& \paperref{\dash} & \paperref{\dash} & \paperref{\dash} & \paperref{\dash} & \paperref{\dash} & \paperref{\dash} \\
{}
& \second{30.908} & \second{0.807} & \best{0.167} & 4.897 & 4.905 & 17 \\
\cmidrule(lr){1-7}

\textbf{EndoPrior-GS (Ours)}
& \best{31.089} & \best{0.856} & \second{0.196}
& \best{2.490} & \best{3.609} & \best{292} \\
\bottomrule
\end{tabular}
\caption{
\textbf{Quantitative comparison on EndoNeRF~\cite{endonerf} and SCARED~\cite{scared}.}
Grey cells denote original-paper values, followed by our reproduced results under a unified protocol. Red superscripts indicate official GitHub links. Best and second-best results are highlighted in red and orange.
}
\label{tab:quantitative_sota}
\vspace{-28pt}
\end{table*}

\begingroup
\renewcommand{\thefootnote}{}
\footnotetext{\scriptsize
\textit{Official implementations:}
\redfn{1} EndoNeRF \href{https://github.com/med-air/EndoNeRF}{GitHub};
\redfn{2} EndoSurf \href{https://github.com/Ruyi-Zha/endosurf}{GitHub};
\redfn{3} Endo-GS \href{https://github.com/HKU-MedAI/EndoGS}{GitHub};
\redfn{4} Endo-4DGS \href{https://github.com/lastbasket/Endo-4DGS}{GitHub};
\redfn{5} EndoGaussian \href{https://github.com/CUHK-AIM-Group/EndoGaussian}{GitHub}.}
\endgroup


Table~\ref{tab:quantitative_sota} compares EndoPrior-GS (Ours) with representative endoscopic reconstruction methods from two categories: (i) neural reconstruction methods, including \textbf{EndoNeRF}~\cite{endonerf} and \textbf{EndoSurf}~\cite{endosurf}; 
and (ii) Gaussian-based methods, including \textbf{EndoGaussian}~\cite{endogaussian}, \textbf{Endo-GS}~\cite{endogs}, and \textbf{Endo-4DGS}~\cite{endo_4dgs}. EndoPrior-GS (Ours) consistently achieves more stable reconstruction in the temporal domain, as reflected by lower Flow Error.
On {EndoNeRF~\cite{endonerf}}: Compared with EndoGaussian~\cite{endogaussian},  EndoPrior-GS (Ours) reduces Flow Error from \(3.791\) to \(2.742\), corresponding to a \textbf{27.7\%} relative reduction, while maintaining competitive rendering quality and speed.
On {SCARED~\cite{scared}}: Compared with Endo-GS~\cite{endogs}, the strongest baseline in terms of Flow Error, EndoPrior-GS (Ours) reduces Flow Error from \(4.867\) to \(3.609\), giving a \textbf{25.8\%} relative reduction. 
Fig.~\ref{fig:final_result} further shows that EndoPrior-GS preserves more continuous tissue structures in challenging regions, consistent with the quantitative improvements.

\FloatBarrier
\subsection{Efficiency and Stability under Matched Initialisation Budgets}
\begin{table}[h]
\centering
\setlength{\tabcolsep}{4.2pt}
\renewcommand{\arraystretch}{0.9}
\small
\resizebox{\linewidth}{!}{%
\begin{tabular}{llrrrrrrr}
\toprule
Budget
& Method
& \(N_{\mathrm{GS}}\)$\downarrow$
& PSNR$\uparrow$
& SSIM$\uparrow$
& LPIPS$\downarrow$
& RMSE$\downarrow$
& Flow Error$\downarrow$
& FPS$\uparrow$ \\
\midrule
10K & EndoGaussian~\cite{endogaussian} & 14.8K & 36.81 & \textbf{0.953} & \textbf{0.084} & 2.97 & 4.46 & 307 \\
& EndoPrior-GS & \textbf{9.7K} & \textbf{36.87} & 0.951 & 0.095 & \textbf{2.47} & \textbf{2.82} & \textbf{310} \\
\midrule
20K & EndoGaussian~\cite{endogaussian} & 22.5K & 37.14 & 0.957 & \textbf{0.066} & 2.80 & 4.11 & 295 \\
& EndoPrior-GS & \textbf{21.6K} & \textbf{38.07} & \textbf{0.967} & 0.071 & \textbf{2.26} & \textbf{2.74} & \textbf{305} \\
\midrule
30K & EndoGaussian~\cite{endogaussian} & 30.7K & 37.32 & 0.959 & \textbf{0.058} & 2.48 & 3.79 & 280 \\
& EndoPrior-GS & \textbf{26.9K} & \textbf{38.12} & \textbf{0.973} & 0.067 & \textbf{2.27} & \textbf{3.07} & \textbf{300} \\
\bottomrule
\end{tabular}%
}
\caption{
\textbf{Efficiency and stability comparison under matched initialisation budgets.}
The budget denotes the nominal initialisation setting, while
\(N_{\mathrm{GS}}\) is the Gaussian count after training.
}
\label{tab:budget_comparison}
\vspace{-20pt}
\end{table}

Table~\ref{tab:budget_comparison} evaluates EndoPrior-GS (Ours) and EndoGaussian~\cite{endogaussian} under matched initialisation budgets. 
This matched-budget analysis decouples the observed temporal stability gain from Gaussian count. Across the 10K, 20K, and 30K initialisation budgets, EndoPrior-GS achieves lower Flow Error with fewer final Gaussians, indicating that the improvement arises independently of increased primitive capacity.

\FloatBarrier
\subsection{Analysis of Temporal Consistency}
\label{sec:temporal_consistency}

\begin{wraptable}[5]{r}{0.49\textwidth}
\vspace{-25pt}
    \centering
    \scriptsize
    \setlength{\tabcolsep}{2.8pt}
    \renewcommand{\arraystretch}{0.95}
    \begin{tabular}{@{}lccc@{}}
        \toprule
        Seq.
        & EndoG.~\cite{endogaussian}
        & EndoPrior-GS
        & Reduction 
        \\
        \midrule
        Pulling
        & 0.4459
        & \textbf{0.3786}
        & \textbf{15.1\%}
        \\
        Cutting
        & 0.1166
        & \textbf{0.1071}
        & \textbf{8.1\%}
        \\
        \bottomrule
    \end{tabular}
    \vspace{-10pt}
    \caption{
    \textbf{Average rendered-depth temporal instability diagnostic.} 
    }
    \label{tab:depth_stability}
\end{wraptable}

\begin{figure}[h]
    \vspace{-15pt}
    \centering
   \includegraphics[width=1.03\linewidth]{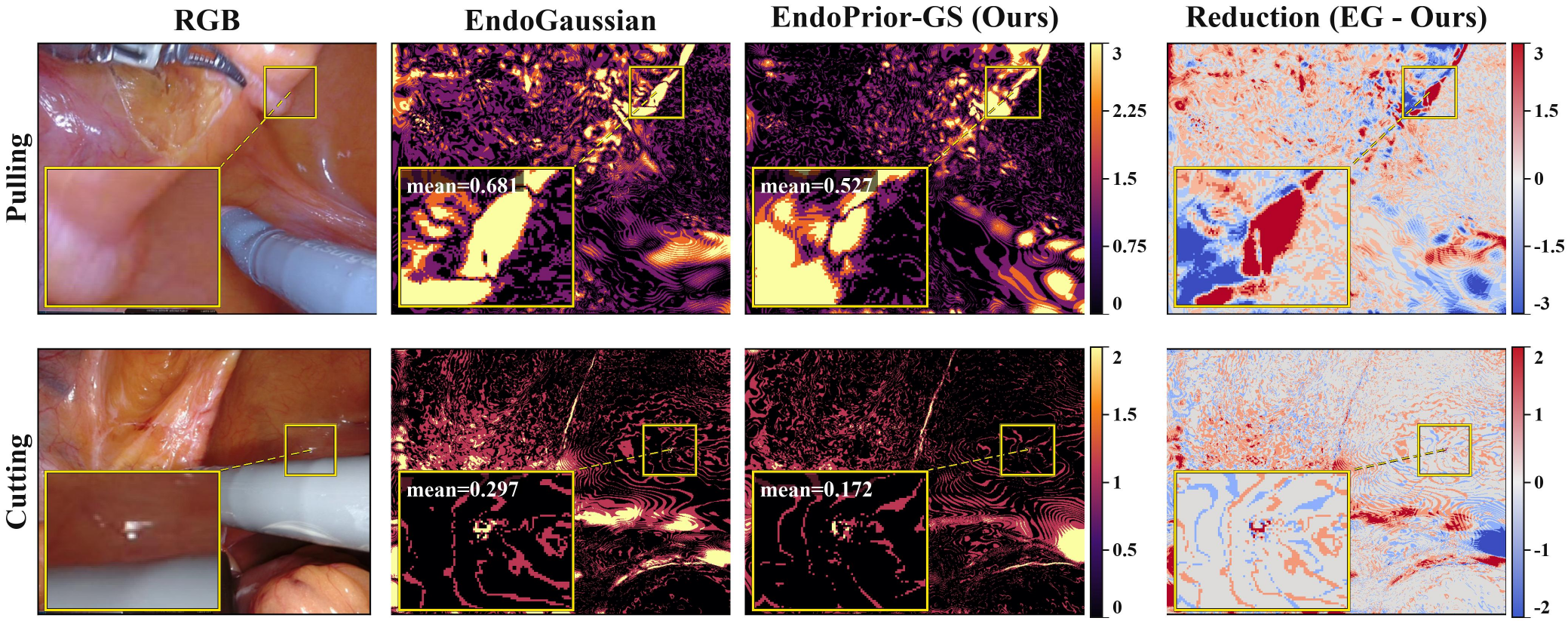}
    \caption{
    \textbf{Comparison of rendered-depth temporal instability diagnostic maps.}
Heat-maps visualise representative pixel-wise adjacent-frame
depth variation.
}
    \label{fig:depth_temporal_stability}
    \vspace{-30pt}
\end{figure}

Beyond the RAFT-based Flow Error, we report the average rendered-depth temporal instability as a complementary diagnostic of temporal stability. We emphasise that these metrics evaluate the consistency of rendered sequences within valid tissue regions.
Table~\ref{tab:depth_stability} reports diagnostic values averaged over all adjacent frame pairs. EndoPrior-GS (Ours) reduces the diagnostic
from \(0.4459\) to \(0.3786\) on \emph{pulling} sequence and from \(0.1166\) to
\(0.1071\) on \emph{cutting} sequence, corresponding to \(15.1\%\) and \(8.1\%\)
reductions, respectively.
Fig.~\ref{fig:depth_temporal_stability}
qualitatively confirms lower local adjacent-frame depth variation in the
highlighted tissue regions. The reduction map is EndoGaussian~\cite{endogaussian} minus EndoPrior-GS (Ours), where positive values indicate lower variation for EndoPrior-GS (Ours). Reported means in Fig.~\ref{fig:depth_temporal_stability} are computed only within the zoomed regions.

\FloatBarrier
\subsection{Analysis of Primitive Distribution}
\label{sec:analysis_allocation}

\begin{table}[h]
\vspace{-12pt}
\centering
\setlength{\tabcolsep}{3.6pt}
\renewcommand{\arraystretch}{1.08}
\scriptsize
\begin{tabular*}{\linewidth}{@{\extracolsep{\fill}}lccccc@{}}
\toprule
Variant
& \(N_{\mathrm{GS}}\downarrow\)
& \(N_{\mathrm{low}\text{-}prior}\)
& \(N_{\mathrm{low}\text{-}\alpha}\downarrow\)
& \(N_{\mathrm{large}\text{-}scale}\downarrow\)
& Flow Error\(\downarrow\) \\
\midrule
Baseline
& 22.5k & 19.9k & 262.0 & 5.0 & 4.112 \\
w/o Texture-aware temp. reg.
& \textbf{21.5k} & 19.9k & 110.3 & 0.7 & 3.152 \\
EndoPrior-GS (Ours)
& 21.6k & 19.9k & \textbf{105.0} & \textbf{0.0} & \textbf{2.742} \\
\bottomrule
\end{tabular*}
\caption{
\textbf{Quantitative primitive count diagnostic analysis.} Counts use fixed post-hoc thresholds after training. The baseline denotes EndoGaussian~\cite{endogaussian} under 20K initialisation budgets.
}
\label{tab:primitive_diagnostics}
\vspace{-20pt}
\end{table}

\begin{figure}[h]
\vspace{-15pt}
\centering
\includegraphics[width=\linewidth]{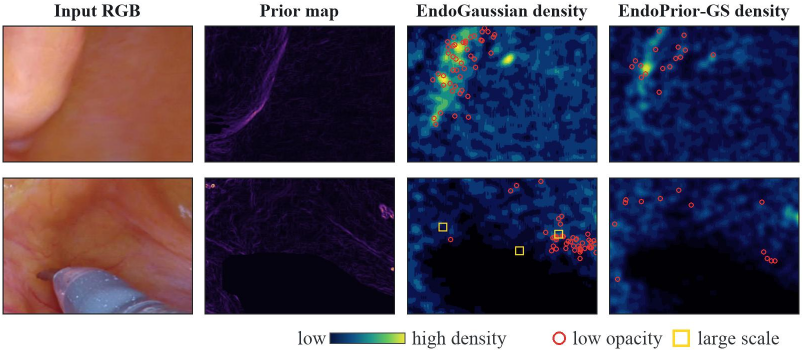}
\caption{
\textbf{Qualitative primitive distribution analysis.}
We show input RGB, joint texture prior \(\rho_t\), and projected Gaussian density, with diagnostic markers computed using the thresholds in Table~\ref{tab:primitive_diagnostics}.
}
\label{fig:allocation_analysis}
\vspace{-10pt}
\end{figure}


The proposed EndoPrior-GS framework suppresses artefact-driven redundant Gaussian growth. 
Table~\ref{tab:primitive_diagnostics} shows that all variants have comparable final Gaussian counts \(N_{\mathrm{GS}}\) (21.5K--22.5K) and identical low-prior primitive counts \(N_{\mathrm{low}\text{-}prior}\) (19.9K), measured with \(\bar{s}_i<0.2\) after at least \(20\) valid observations. Under these comparable counts, EndoPrior-GS (Ours) reduces low-opacity primitives \(N_{\mathrm{low}\text{-}\alpha}\) with opacity \(<0.01\) from 262.0 to 105.0, and large-scale primitives with normalised-coordinate scale \(>40\) from 5.0 to 0.0.
These diagnostics indicate that the proposed prior-guided strategy does not simply remove low-prior primitives. Instead, it suppresses opacity-ineffective and
over-scaled primitive growth while maintaining comparable projected primitive coverage over valid tissue regions.
Fig.~\ref{fig:allocation_analysis} explains this behaviour qualitatively. Compared with EndoGaussian~\cite{endogaussian}, EndoPrior-GS reduces redundant and unstable growth around surgical-tool and tissue boundary areas.

\FloatBarrier
\subsection{Ablation Study}

\paragraph{Texture-aware temporal regularisation ablation.}
Table~\ref{tab:temp_reg} isolates the texture-aware temporal regularisation term in Sec.~\ref{sec:reg}.
Prior-guided weighting substantially reduces Flow Error from
3.569 to 3.003 (15.9\%) under identical temporal constraints.
Distance control provides an additional gain, while the
full formulation achieves the lowest RMSE with comparable PSNR.

\begin{table}[h]
\vspace{-8pt}
\centering
\small
\setlength{\tabcolsep}{3pt}
\renewcommand{\arraystretch}{1.1}
\definecolor{paperrefgrey}{RGB}{242,242,242}
\newcommand{\best}[1]{\cellcolor{red!18}\textbf{#1}}
\newcommand{\second}[1]{\cellcolor{orange!22}\textbf{#1}}
\resizebox{\linewidth}{!}{%
\begin{tabular}{@{}lcccccccc@{}}
\toprule
Variant
& Pair. w.
& Disp. c.
& Dist. c.
& PSNR$\uparrow$
& SSIM$\uparrow$
& LPIPS$\downarrow$
& RMSE$\downarrow$
& Flow Error$\downarrow$ \\
\midrule

Baseline
& $\times$ & $\times$ & $\times$
& \textbf{37.147}
& 0.957
& \textbf{0.066}
& 2.804
& 4.112 \\

w/o Pairwise weighting
& $\times$ & $\checkmark$ & $\checkmark$
& 37.017 
& 0.959
& 0.068
& 2.785
& 3.569 \\

w/o Distance control
& $\checkmark$ & $\checkmark$ & $\times$
& 37.104
& 0.959
& 0.070
& 2.797
& 3.144 \\

w/ Temporal reg.
& $\checkmark$ & $\checkmark$ & $\checkmark$
& 37.129
& \textbf{0.961}
& {0.069}
& \textbf{2.575}
& \textbf{3.003} \\

\bottomrule
\end{tabular}%
}
\caption{
\textbf{Ablation of texture-aware temporal regularisation.}
Pair. w.: pairwise affinity weighting;
Disp. c.: primitive displacement control;
Dist. c.: pairwise distance control.
}
\label{tab:temp_reg}
\vspace{-18pt}
\end{table}

\paragraph{Components ablation.}
Table~\ref{tab:component_ablation} evaluates progressive component
contributions under the matched 20K initialisation budget.
Prior-guided initialisation reduces Flow Error from 4.112
to 3.343, and prior-guided density
control further lowers it to 3.152.
Temporal regularisation constrains deformed Gaussian centres
to discourage extreme local deformation, further reducing
Flow Error to 2.701 while achieving the highest PSNR and
lowest RMSE among the tested variants.

\begin{table}[h]
\centering
\small
\setlength{\tabcolsep}{3pt}
\renewcommand{\arraystretch}{1.1}
\resizebox{\linewidth}{!}{%
\begin{tabular}{@{}lcccccccc@{}}
\toprule
Variant
& Pri. init.
& Pri. dens.
& Temp. reg.
& PSNR$\uparrow$
& SSIM$\uparrow$
& LPIPS$\downarrow$
& RMSE$\downarrow$
& Flow Error$\downarrow$ \\
\midrule

Prior init. 
& $\checkmark$ & $\times$ & $\times$
& 37.080
& 0.959
& \textbf{0.060}
& 2.727
& 3.343 \\

+ Prior dens.
& $\checkmark$ & $\checkmark$ & $\times$
& 37.165
& 0.961
& 0.065
& 2.479
& 3.152 \\

Full model
& $\checkmark$ & $\checkmark$ & $\checkmark$
& \textbf{38.041}
& \textbf{0.967}
& 0.071
& \textbf{2.260}
& \textbf{2.701} \\

\bottomrule
\end{tabular}%
}
\caption{
\textbf{Ablation study of the components.} All variants are evaluated under the
matched 20K budget.
}
\label{tab:component_ablation}
\end{table}

\vspace{-20pt}

\section{Conclusion}
\label{sec:conclusion}
This paper presents EndoPrior-GS, a novel dynamic endoscopic Gaussian reconstruction framework that uses a heuristic and training-free joint texture prior to guide Gaussian primitive evolution. 
We further extend the texture prior to the temporal domain through a texture-aware term that dynamically weighs pairwise primitive contribution during training. 
Extensive experiments show that EndoPrior-GS reduces Flow Error by 27.7\% and 25.8\% on EndoNeRF and SCARED, respectively. 
For temporal evaluation, Flow Error is computed from RAFT-estimated image-plane optical flow, as dense 3D tissue-motion ground truth is unavailable in current benchmarks; we therefore use it together with an adjacent-frame rendered-depth diagnostic as supporting evidence for rendered temporal stability and consistency.
The current prior is derived from low-level image cues and assumes available tool-filtered valid tissue support. 
It may become less discriminative under pronounced blur, over-saturation, dense smoke occlusion, highly ambiguous tissue texture, or noisy predicted masks.

%
%
\clearpage
\FloatBarrier
\bibliographystyle{splncs04}
\bibliography{refs}

@String(ICIP  = {IEEE Int. Conf. Image Process.})

@String(ICIP  = {ICIP})

@Article{3dgs,
      author       = {Kerbl, Bernhard and Kopanas, Georgios and Leimk{\"u}hler, Thomas and Drettakis, George},
      title        = {{3D Gaussian Splatting} for Real-Time Radiance Field Rendering},
      journal      = {ACM Transactions on Graphics},
      number       = {4},
      volume       = {42},
      month        = {July},
      year         = {2023}}

@article{endogaussian,
  title={Foundation model-guided {Gaussian Splatting} for {4D} reconstruction of deformable tissues},
  author={Liu, Yifan and Li, Chenxin and Liu, Hengyu and Yang, Chen and Yuan, Yixuan},
  journal={IEEE Transactions on Medical Imaging},
  year={2025},
  publisher={IEEE}
}

@inproceedings{endonerf,
  title={Neural rendering for {Stereo 3D Reconstruction} of deformable tissues in robotic surgery},
  author={Wang, Yuehao and Long, Yonghao and Fan, Siu Hin and Dou, Qi},
  booktitle={International conference on medical image computing and computer-assisted intervention},
  pages={431--441},
  year={2022},
  organization={Springer}
}

@inproceedings{endogs,
  title={{EndoGS}: Deformable endoscopic tissues reconstruction with {Gaussian Splatting}},
  author={Zhu, Lingting and Wang, Zhao and Cui, Jiahao and Jin, Zhenchao and Lin, Guying and Yu, Lequan},
  booktitle={International Conference on Medical Image Computing and Computer-Assisted Intervention},
  pages={135--145},
  year={2024},
  organization={Springer}
}

@inproceedings{sfm,
  title={{Structure-from-Motion} revisited},
  author={{Schonberger}, Johannes L and Frahm, Jan-Michael},
  booktitle={Proceedings of the IEEE conference on computer vision and pattern recognition},
  pages={4104--4113},
  year={2016}
}

@article{nerf,
  title={{NeRF}: representing scenes as neural radiance fields for view synthesis},
  author={Mildenhall, Ben and Srinivasan, Pratul P and Tancik, Matthew and Barron, Jonathan T and Ramamoorthi, Ravi and Ng, Ren},
  journal={Communications of the ACM},
  volume={65},
  number={1},
  pages={99--106},
  year={2021},
  publisher={ACM New York, NY, USA}
}

@article{scared,
  title={Stereo correspondence and reconstruction of endoscopic data challenge},
  author={Allan, Max and {McLeod}, Jonathan and Wang, Congcong and Rosenthal, Jean Claude and Hu, Zhenglei and Gard, Niklas and Eisert, Peter and Fu, Ke Xue and Zeffiro, Trevor and Xia, Wenyao and others},
  journal={arXiv preprint arXiv:2101.01133},
  year={2021}
}

@inproceedings{endosurf,
  title={{EndoSurf}: Neural surface reconstruction of deformable tissues with stereo endoscope videos},
  author={Zha, Ruyi and Cheng, Xuelian and Li, Hongdong and Harandi, Mehrtash and Ge, Zongyuan},
  booktitle={International conference on medical image computing and computer-assisted intervention},
  pages={13--23},
  year={2023},
  organization={Springer}
}

@article{3d_reconstruction_surgical,
  title={{3D} renal model for surgical planning of partial nephrectomy: a way to improve surgical outcomes},
  author={Bianchi, Lorenzo and Cercenelli, Laura and Bortolani, Barbara and Piazza, Pietro and Droghetti, Matteo and Boschi, Sara and Gaudiano, Caterina and Carpani, Giulia and Chessa, Francesco and Lodi, Simone and others},
  journal={Frontiers in {Oncology}},
  volume={12},
  pages={1046505},
  year={2022},
  publisher={Frontiers Media SA}
}

@article{3d_reconstruction_surgical_train,
  title={Virtual reality training improves operating room performance: results of a randomized, double-blinded study},
  author={Seymour, Neal E and Gallagher, Anthony G and Roman, Sanziana A and {O’Brien}, Michael K and Bansal, Vipin K and Andersen, Dana K and Satava, Richard M},
  journal={Annals of surgery},
  volume={236},
  number={4},
  pages={458--464},
  year={2002},
  publisher={LWW}
}

@article{3d_reconstruction_surgical_train_need,
  title={Virtual reality training compared with apprenticeship training in laparoscopic surgery: a meta-analysis},
  author={Portelli, M and Bianco, SF and Bezzina, T and Abela, JE},
  journal={The Annals of the Royal College of Surgeons of England},
  volume={102},
  number={9},
  pages={672--684},
  year={2020},
  publisher={Royal College of Surgeons}
}

@article{surgical_robots,
  title={Levels of Autonomy in {FDA}-Cleared Surgical Robots: A Systematic Review},
  author={Lee, Audrey and Baker, Turner S and Bederson, Joshua B and Rapoport, Benjamin I},
  journal={NPJ Digital Medicine},
  volume={7},
  number={1},
  pages={103},
  year={2024},
  publisher={Nature Publishing Group UK London}
}

@inproceedings{endo_4dgx,
  title={{Endo-4DGX}: Robust endoscopic scene reconstruction and illumination correction with {Gaussian Splatting}},
  author={Huang, Yiming and Bai, Long and Cui, Beilei and Li, Yanheng and Chen, Tong and Wang, Jie and Wu, Jinlin and Lei, Zhen and Liu, Hongbin and Ren, Hongliang},
  booktitle={International Conference on Medical Image Computing and Computer-Assisted Intervention},
  pages={181--191},
  year={2025},
  organization={Springer}
}

@inproceedings{endo_4dgs,
  title={{Endo-4DGS}: Endoscopic monocular scene reconstruction with {4D} {Gaussian Splatting}},
  author={Huang, Yiming and Cui, Beilei and Bai, Long and Guo, Ziqi and Xu, Mengya and Islam, Mobarakol and Ren, Hongliang},
  booktitle={International Conference on Medical Image Computing and Computer-Assisted Intervention},
  pages={197--207},
  year={2024},
  organization={Springer}
}

@inproceedings{e_dssr,
  title={{E-DSSR}: efficient dynamic surgical scene reconstruction with transformer-based stereoscopic depth perception},
  author={Long, Yonghao and Li, Zhaoshuo and Yee, Chi Hang and Ng, Chi Fai and Taylor, Russell H and Unberath, Mathias and Dou, Qi},
  booktitle={International Conference on Medical Image Computing and Computer-Assisted Intervention},
  pages={415--425},
  year={2021},
  organization={Springer}
}

@inproceedings{raft,
  title={{RAFT}: Recurrent All-Pairs Field Transforms for Optical Flow},
  author={Teed, Zachary and Deng, Jia},
  booktitle={European conference on computer vision},
  pages={402--419},
  year={2020},
  organization={Springer}
}

@article{serv_ct,
  title={{SERV-CT}: A disparity dataset from cone-beam {CT} for validation of endoscopic {3D} reconstruction},
  author={Edwards, PJ Eddie and Psychogyios, Dimitris and Speidel, Stefanie and Maier-Hein, Lena and Stoyanov, Danail},
  journal={Medical image analysis},
  volume={76},
  pages={102302},
  year={2022},
  publisher={Elsevier}
}

@inproceedings{pixelgs,
  title={{Pixel-GS}: Density Control with Pixel-Aware Gradient for {3D} {Gaussian Splatting}},
  author={Zhang, Zheng and Hu, Wenbo and Lao, Yixing and He, Tong and Zhao, Hengshuang},
  booktitle={European Conference on Computer Vision},
  pages={326--342},
  year={2024},
  organization={Springer}
}

@inproceedings{hsv,
  title={Generic and real-time detection of specular reflections in images},
  author={Morgand, Alexandre and Tamaazousti, Mohamed},
  booktitle={2014 International conference on computer vision theory and applications (VISAPP)},
  volume={1},
  pages={274--282},
  year={2014},
  organization={IEEE}
}

@inproceedings{specular,
  title={Adaptive specular reflection detection and inpainting in colonoscopy video frames},
  author={Akbari, Mojtaba and Mohrekesh, Majid and Najariani, Kayvan and Karimi, Nader and Samavi, Shadrokh and Soroushmehr, SM Reza},
  booktitle={2018 25th IEEE international conference on image processing ({ICIP})},
  pages={3134--3138},
  year={2018},
  organization={IEEE}
}

@article{huber,
  title={Robust Estimation of a Location Parameter},
  author={Huber, Peter J.},
  journal={The Annals of Mathematical Statistics},
  volume={35},
  number={1},
  pages={73--101},
  year={1964}
}

@inproceedings{surgicalgaussian,
  title={{SurgicalGaussian}: Deformable {3D Gaussians} for high-fidelity surgical scene reconstruction},
  author={Xie, Weixing and Yao, Junfeng and Cao, Xianpeng and Lin, Qiqin and Tang, Zerui and Dong, Xiao and Guo, Xiaohu},
  booktitle={International Conference on Medical Image Computing and Computer-Assisted Intervention},
  pages={617--627},
  year={2024},
  organization={Springer}
}

@inproceedings{surgical_gaussian_surfels,
  title={{Surgical Gaussian Surfels}: Highly accurate real-time surgical scene rendering using {Gaussian Surfels}},
  author={Sunmola, Idris O and Zhao, Zhenjun and Schmidgall, Samuel and Wang, Yumeng and Scheikl, Paul Maria and Pham, Viet and Krieger, Axel},
  booktitle={Proceedings of the IEEE/CVF Winter Conference on Applications of Computer Vision},
  pages={4515--4524},
  year={2026}
}

@inproceedings{free_surgs,
  title={{Free-SurGS}: {SfM}-Free {3D Gaussian Splatting} for Surgical Scene Reconstruction},
  author={Guo, Jiaxin and Wang, Jiangliu and Kang, Di and Dong, Wenzhen and Wang, Wenting and Liu, Yun-hui},
  booktitle={International Conference on Medical Image Computing and Computer-Assisted Intervention},
  pages={350--360},
  year={2024},
  organization={Springer}
}

@inproceedings{pixelsplat,
  title={{pixelSplat}: {3D} {Gaussian} Splats from Image Pairs for Scalable Generalizable {3D} Reconstruction},
  author={Charatan, David and Li, Sizhe Lester and Tagliasacchi, Andrea and Sitzmann, Vincent},
  booktitle={Proceedings of the IEEE/CVF conference on computer vision and pattern recognition},
  pages={19457--19467},
  year={2024}
}

@inproceedings{splatter_image,
  title={{Splatter Image}: Ultra-fast single-view {3D} reconstruction},
  author={Szymanowicz, Stanislaw and Rupprecht, Chrisitian and Vedaldi, Andrea},
  booktitle={Proceedings of the IEEE/CVF conference on computer vision and pattern recognition},
  pages={10208--10217},
  year={2024}
}

@inproceedings{mvsplat,
  title={{MVSplat}: Efficient {3D Gaussian Splatting} from Sparse Multi-view Images},
  author={Chen, Yuedong and Xu, Haofei and Zheng, Chuanxia and Zhuang, Bohan and Pollefeys, Marc and Geiger, Andreas and Cham, Tat-Jen and Cai, Jianfei},
  booktitle={European conference on computer vision},
  pages={370--386},
  year={2024},
  organization={Springer}
}

@inproceedings{depthsplat,
  title={{DepthSplat}: Connecting {Gaussian Splatting} and depth},
  author={Xu, Haofei and Peng, Songyou and Wang, Fangjinhua and Blum, Hermann and Barath, Daniel and Geiger, Andreas and Pollefeys, Marc},
  booktitle={Proceedings of the Computer Vision and Pattern Recognition Conference},
  pages={16453--16463},
  year={2025}
}

@article{ssim,
  title={Image quality assessment: from error visibility to structural similarity},
  author={Wang, Zhou and Bovik, Alan C and Sheikh, Hamid R and Simoncelli, Eero P},
  journal={{IEEE Transactions on Image Processing}},
  volume={13},
  number={4},
  pages={600--612},
  year={2004},
  publisher={IEEE}
}

@inproceedings{lpips,
  title={The unreasonable effectiveness of deep features as a perceptual metric},
  author={Zhang, Richard and Isola, Phillip and Efros, Alexei A and Shechtman, Eli and Wang, Oliver},
  booktitle={Proceedings of the IEEE conference on computer vision and pattern recognition},
  pages={586--595},
  year={2018}
}
\end{document}